# Research on Data Fusion Algorithm Based on Deep Learning in Target Tracking

Huihui Wu [1] *, Jianquan Li [1], Peijun Zhang [1]

[1] School of Computer Science, Xijing University, No.1 Xijing Road, Xi'an 710071, China

*Correspondence information:
Full name: Huihui Wu, Affiliation: School of Computer Science, XiJing University,
Detailed permanent address: 135 room, Teacher Development Center, No.1 Xijing Road, Xi'an 710071, China.
Email address: wuhh723@163.com,
Telephone number: +86-180-6687-0412

**Abstract.** Aiming at the limitation that deep long and short-term memory network (DLSTM) algorithm cannot perform parallel computing and cannot obtain global information, in this paper, feature extraction and feature processing are firstly carried out according to the characteristics of eye movement data and tracking data, then by introducing a convolutional neural network (CNN) into a deep long and short-term memory network, developed a new network structure and designed a fusion strategy, an eye tracking data fusion algorithm based on long and short-term memory network is proposed. The experimental results show that compared with the two fusion algorithms based on deep learning, the algorithm proposed in this paper performs well in terms of fusion quality.

# 1. Introduction

At present, Kernelized Correlation Filter (KCF) is an online learning-based target tracking algorithm [3] for real-time tracking of moving objects, which cleverly uses cyclic shifts to generate training samples, bridging the gap between traditional digital signal processing [19] and machine learning [2]. But when the object encounters a special situation, such as partial occlusion [8], the position and shape of the target changes instantaneously [5], severely uneven light intensity [10], the tracking effect of the KCF algorithm in the target is not ideal, and even lost [1]. In addition, gaze target tracking is also a method for real-time tracking of moving objects [7], which adopts a model-based gaze solution algorithm [6], but it has high requirements for accuracy and is easily affected by people's glasses, emotions and the surrounding environment [12], and it is extremely sensitive to the viewing angle, eye image processing, light source and sensor placement [4], so that the tracking of the target is not very accurate. When using these two methods to track the same moving object, how to fuse the tracking data generated by KCF and the eye movement data generated by gaze target tracking to make the target tracking result more accurate is an urgent problem to be solved.

Gao proposed a fusion algorithm based on a deep learning architecture, which uses a linear combination of two Convolutional Neural Networks to achieve the final fusion purpose [13]. Daniel proposed a Deep Convolutional and Long Short Term Memory Recurrent Neural Networks (DeepConvLSTM) network structure [15], which adopts convolutional layer, dence layer, softmax layer, and regards the whole network structure as a fusion process. Yao proposed a Deep Learning Framework for Time-Series Mobile Sensing Data Processing (DeepSense) network structure [16], which takes the entire network as a fusion process. By using Visual Geometry Group (VGG) [9] and Long Short-term Memory (LSTM) [11] network to extract the spatial and temporal characteristics of sample data. Li proposed a multi-modal VGG-LSTM network [17] structure and designed a fusion strategy, which takes the fully connected layer [14] as the final fusion result. Alaa proposed a Deep Long and Short-term Memory algorithm [18], which uses a genetic algorithm to initialize various parameters to achieve fusion by stacking 3-layer LSTM architectures.

Constrained by conditions, commonly used fusion algorithms are difficult to accurately fuse eye movement data and tracking data. Therefore, according to the structural characteristics of eye movement data and tracking data, improving the existing data fusion algorithm is the technical key to effectively process eye movement data and tracking data, so that it can be more fully fused to achieve accurate tracking of targets in special circumstances. This paper improves the DLSTM algorithm according to the characteristics of eye movement and tracking data, and proposes a new eye tracking data fusion algorithm based on LSTM network, denoted as EyeLSTM algorithm, which combines a CNN network with a LSTM network, a new network structure is developed and a data fusion strategy is designed to achieve accurate target tracking.

# 2. EyeLSTM algorithm

*2.1 Algorithmic ideas*

The DLSTM algorithm forms a deep LSTM network by stacking multiple LSTM blocks using a deep recurrent network, the network structure of DLSTM is shown in Fig. 1.

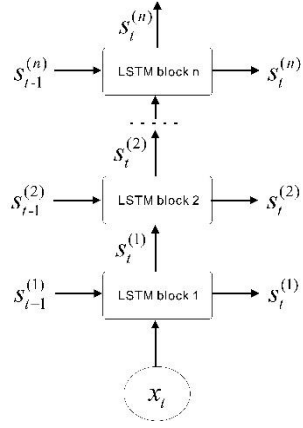

**Fig. 1** Network structure of DLSTM

Although the stacked architecture of DLSTM maintains the advantages of LSTM in processing long-term dependent time series sequences, it cannot obtain global information at one time and cannot perform parallel operations, resulting in long training time and low efficiency. Furthermore, the robustness of the system is not good for time series with different properties. Therefore, this paper improves the network structure of DLSTM. Here, a CNN deep learning network is introduced into LSTM, which can extract features from the input data set and capture global information, and can perform parallel operations, but it has no time relationship, each data in the data set is regarded as independent of each other. Although CNN can extract features, it still needs to manually extract features in advance, because the features extracted by CNN are relatively abstract and the extracted features are not comprehensive. Then combine CNN and LSTM to obtain a network structure, input the eye movement data of the dataset into the network, and after 100 repeated experiments, the optimal network parameters can be obtained, that is a new network structure. In the same way, a new network structure is trained using the tracking data of the dataset, and finally a data fusion strategy is designed based on the two network structures, the results obtained by each network are linearly combined to obtain a final data fusion result. It is represented by formula (1), where the weight coefficient is selected according to the Softmax algorithm. Based on the actual situation of this paper, the Softmax algorithm [20] can be used to obtain $w_1 = 0.5$, $w_2 = 0.5$.

$$fusion = w_1 EyeData_{EyeLSTM-1} + w_2 TrackData_{EyeLSTM-2}, \qquad (1)$$

Where *fusion* represents the fusion result, $EyeData_{EyeLSTM-1}$ represents a result obtained by eye movement data trained by the EyeLSTM network, $TrackData_{EyeLSTM-2}$ represents a result obtained by training the EyeLSTM network on the tracking data.

The network structure of EyeLSTM is shown in Fig. 2.

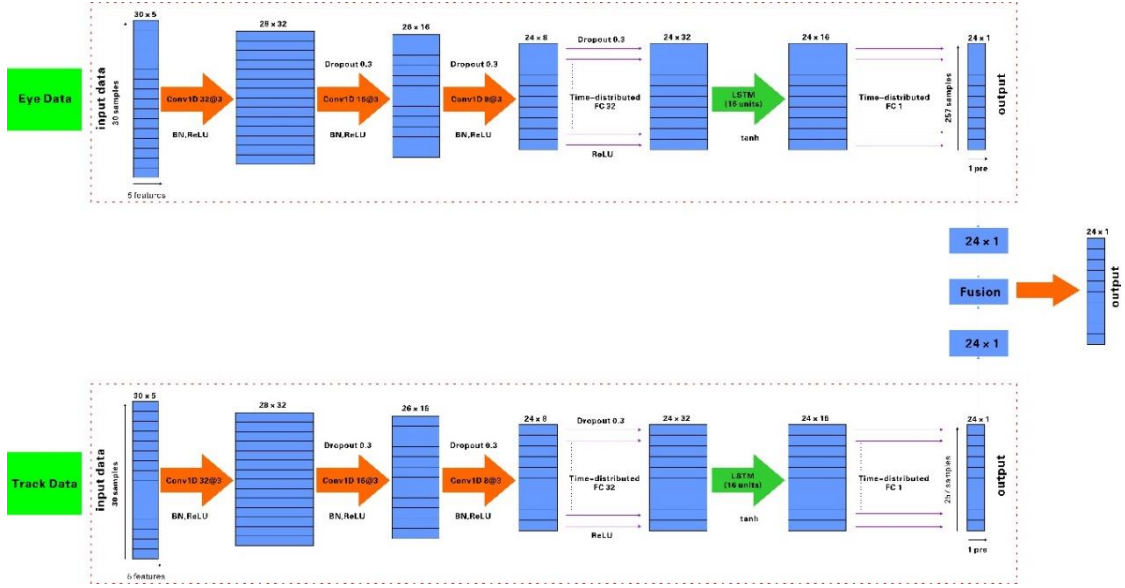

**Fig. 2** Network structure of EyeLSTM

*2.2 Algorithmic steps*

The steps of the EyeLSTM algorithm proposed in this paper are as follows:

**Input:** eye movement dataset $X$, tracking dataset $Y$.

**Output:** the optimal fusion result *fusion*.

**Step1:** Extract features from eye movement datasets $X$ and tracking datasets $Y$ and perform feature processing.

**Step2:** Group the processed eye movement dataset $\hat{X}$ and the tracking dataset $\hat{Y}$ respectively. Taking each 24 data as a group, we can get $\hat{X} = (x_1, x_2, ..., x_n)$, $\hat{Y} = (y_1, y_2, ..., y_m)$, where $x_i = (x_{i1}, x_{i2}, ..., x_{i24})$, $i = 1, 2, ..., n$, $y_j = (y_{j1}, y_{j2}, ..., y_{j24})$, $j = 1, 2, ..., m$.

**Step3:** $t = 1$.

**Step4:** Mirror filling of $x_t$ and $y_t$ respectively to expand it to 30 data, that is $x'_t = (x'_{t1}, x'_{t2}, ..., x'_{t30})$, $y'_t = (y'_{t1}, y'_{t2}, ..., y'_{t30})$.

**Step5:** Perform three layers of convolution on $x'_t$ and $y'_t$ respectively, one layer of full connection operation, one-layer LSTM network and one-layer full-connection operation on and respectively to obtain $x^*_t = (x^*_{t1}, x^*_{t2}, ..., x^*_{t24})$, $y^*_t = (y^*_{t1}, y^*_{t2}, ..., y^*_{t24})$.

**Step6:** The weighted average of $x^*_t$ and $y^*_t$ can get the fusion result of the first group $fusion_t$, where $fusion_t = (fusion_{t1}, fusion_{t2}, ..., fusion_{t24})$.

**Step7:** When $t > \max(n, m)$, the iteration stops, and all fusion results *fusion* can be obtained, where $fusion = (fusion_1, fusion_2, ..., fusion_t)$.

**3. Data set**

In order to verify the performance of the algorithm proposed in this paper, this paper selects four groups of video sequences in the Visual Tracker Benchmark, namely Bird1, Bolt, ClifBar, MotorRolling, and uses their groundtruth values as labels. These four video sequences have attributes such as illumination, partial occlusion, and deformation. Then, the eye tracking system (eye tracker) and KCF algorithm are used to collect the corresponding eye movement and tracking data respectively, so as to realize the fusion of eye movement and tracking data.

*3.1 Data collection*

First, the eye tracking system is used to collect the eye movement data of the experimenter. The system mainly consists of 2 parts: an eye tracker and a laptop. Before the experiment starts, each experimenter needs to perform a calibration process: the experimenter needs to be about 60cm away from the computer screen. First, look at a water-shaped ball on the screen to calibrate the eye tracker, and then look at a point in the center of the screen and 5 calibration points around it. At this time, the eye tracker will capture the data of the eye movement trajectory, and calculate the position of the experimenter's gaze point according to its built-in calibration algorithm. Due to the large amount of data and space limitations, this paper only shows the eye movement data of the Bird1 video sequence. The motion positions of some objects in the Bird1 video sequence are shown in Fig. 3.

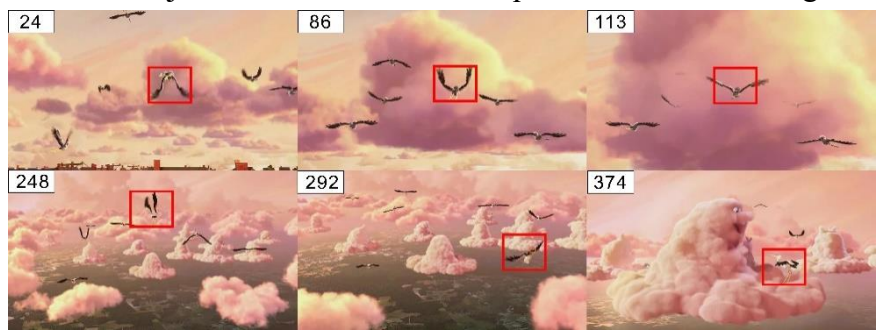

**Fig. 3** Movement positions of some objects in the Bird1 sequence

In order to more intuitively observe the eye movement data in Fig. 3 collected by the eye tracker, Fig. 4 shows the distribution of Bird1 video sequences.

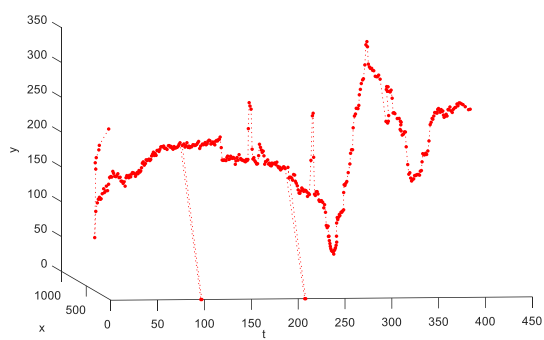

**Fig. 4** The human eye tracks the motion trajectory of the Bird1 sequence target frame

Secondly, the KCF algorithm is used to track the targets of 8 video sequences, and a series of tracking data are obtained. Due to the large amount of data and space limitations, this paper only shows the tracking data of MontorRolling video sequences. The motion positions of some objects in the MontorRolling video sequence are shown in Fig. 5.

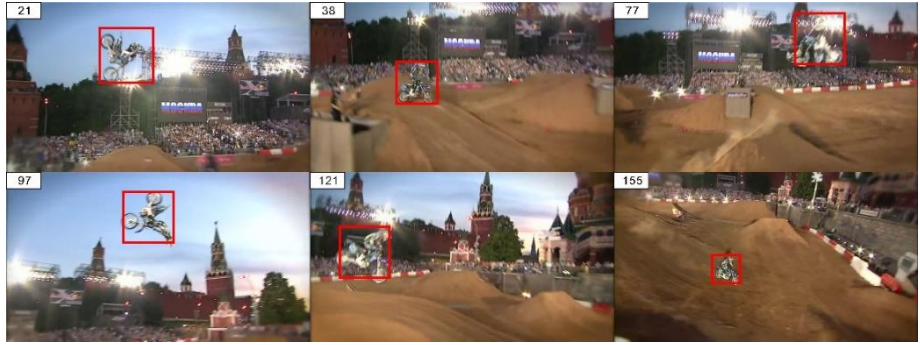

**Fig. 5** Movement positions of some objects in the MotorRolling sequence

To more intuitively observe the tracking data in Fig. 5 collected by the KCF algorithm, Fig. 6 shows the distribution of the MontorRolling video sequence.

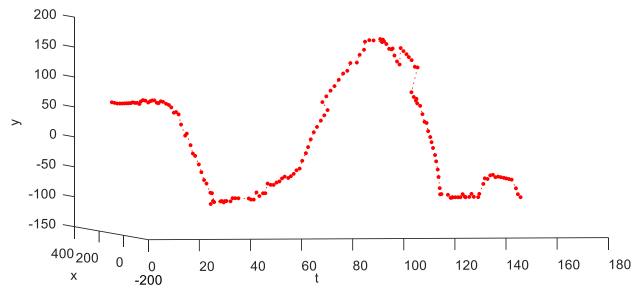

**Fig. 6** Tracking data in MotorRolling sequence

*3.2 Feature extraction*

In the collection process of eye movement data, since the time interval of playing each picture is about 70ms, and the time interval of eye movement data collected by the eye tracker is about 18ms. Therefore, multiple sets of eye movement data can be collected during the playback time of each frame of picture, that is, each frame of picture corresponds to multiple eye movement data. However, the target and background of each picture are fixed, and most of the eye movement data collected in this picture will appear smooth trailing, that is, the fluctuation range of multiple eye movement data corresponding to the same frame is not large, so that the amount of information in the data is too large. It is necessary to extract features from this data to fully reflect its structural characteristics.

Since the eye movement data corresponding to each picture does not change much. Therefore, this paper performs a weighted average of all the eye movement data corresponding to each picture, and uses the result as the feature of the set of data, which is larger than the smallest data in this set of data, smaller than the largest data, and in the middle. The overall distribution structure of the data can be maintained, so it is reasonable and feasible to use the weighted average method to extract features from the eye movement data. In addition, the image pixels of each dataset are different. For the Bird1 dataset, the original image size is 720*400 pixels, which accounts for about half of the entire computer screen, and the target is a smaller object in the original image. If an eye tracker is used to collect the eye movement data of the target in this picture, because the target is too small, the human eye is prone to fatigue when gazing at the target, so the collected eye movement data is inaccurate. Therefore, in the process of collecting eye movement data, the picture is enlarged to 1440*900 pixels, and then collected by the eye tracker. The eye movement data finally obtained is collected after the pixels of the picture are enlarged. In order to keep the coordinate system consistent with the original image, the value of the $x$-coordinate corresponding to its spatial position needs to be reduced to 2 times, and the value of the $y$-coordinate is reduced to 2.25 times, and the extracted value is the real feature of the eye movement data.

*3.3 Feature processing*

The data collected by the eye tracker will lose frames. The main reasons are: (1) in the process of collecting eye movement data, the eye tracker may not be able to collect data at these moments due to the experimenter's blinking, inattention or head movement; (2) The properties of the eye tracker itself have an impact on the acquisition process. If the sampling rate is below 60Hz, the data processing speed is too slow and there will be frame loss. In addition, the original eye movement data contains a lot of noise, which has a great impact on the fusion accuracy. In order to deal with these missing frames and filter out these noises, before fusion, this paper first uses heuristic filtering algorithm to denoise the dataset to make its spatial distribution smoother. According to the structural characteristics of the data, the heuristic filtering algorithm designs a two-stage filter through the "rule of thumb", that is, compares each data point with its adjacent data points, and appropriately corrects the value of the data point according to the comparison results to obtain. It makes the overall distribution of the data look smoother. Fig. 7 is a comparison of eye movement data before and after filtering:

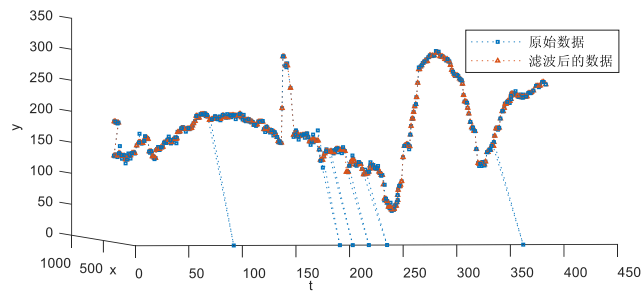

**Fig. 7** Comparison of eye movement data before and after filtering in Bird1 sequence

As can be seen from Fig. 7, after the heuristic filtering algorithm is used to process the eye movement data, the position of the fixation point becomes more compact, and the smooth trailing point is also more continuous and smooth. Therefore, the filtered data is more in line with the real movement of the human eye in the process of gazing at the target.

**4. Experimental results and analysis**

There are many evaluation indicators for data fusion. This paper selects the commonly used indicators for evaluating fusion performance, which are root mean square error (RMSE), root mean square percentage error (RMSPE), mean absolute error (MAE), Mean Absolute Percentage Error (MAPE).

In order to test the fusion performance of the EyeLSTM algorithm, considering the degree of eye fatigue of subjects who have been staring at a certain target, this paper selects four groups of video sequences in the Visual Tracker Benchmark, namely Bird1, Bolt, ClifBar, and MotorRolling. The continuous playback time of these four groups of video sequences is about 2 minutes, and after each video sequence is played, the eyes of the subjects can rest for a while. These four sets of video sequences are used to collect eye movement and tracking data, including 10 sets of eye movement data and 10 sets of tracking data. Two deep learning algorithms are compared, and all experiments are run in the environment of Ubuntu 18.4.

Since the goal of these four video sequences is to move the front, back, up, down, left, and right positions on a plane. The forward and backward movement will cause the overlapping of the target coordinates in the plane. Therefore, in order to better demonstrate the effect of fusion, this paper adds a time axis in Fig. 8-9 to form three-dimensional coordinates.

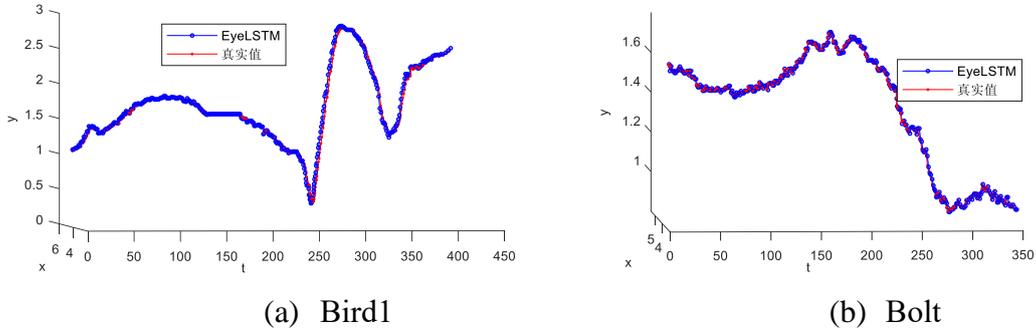

(a)　Bird1　　　　　　　　　　　(b)　Bolt

**Fig. 8** Fusion results of EyeLSTM algorithm on Bird1 and Bolt datasets

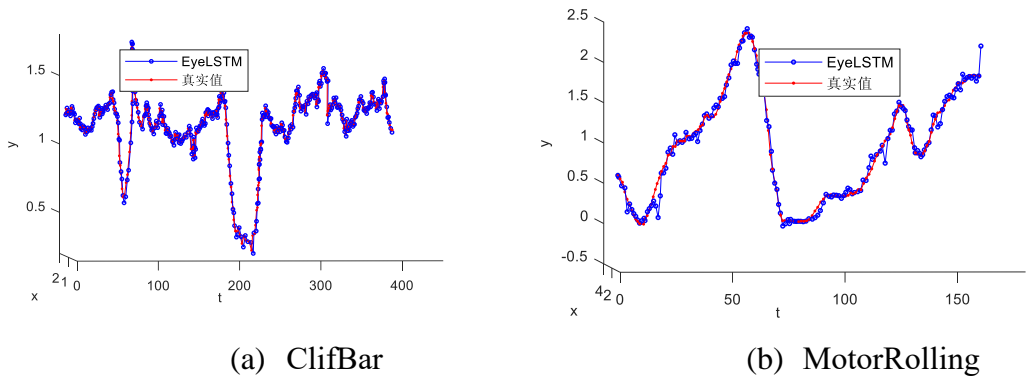

(a)　ClifBar　　　　　　　　　　(b)　MotorRolling

**Fig. 9** Fusion results of the EyeLSTM algorithm on the ClifBar and MotorRolling datasets

It can be seen from Fig. 8-9 that the fusion results of the EyeLSTM algorithm on most of the data sets are closer to the true value of the target, indicating that the data fusion strategy designed by the algorithm in this paper is effective.

In order to comprehensively evaluate the fusion performance of the EyeLSTM algorithm, this paper compares the EyeLSTM algorithm with the MLP and DLSTM algorithms. Table 1 shows the index values of the four data sets obtained by the data fusion of EyeLSTM and the other two fusion algorithms.

Table 1. Fusion results of three fusion algorithms on four datasets

| Algorithm | Bird1 | | | | Bolt | | | |
|---|---|---|---|---|---|---|---|---|
| | RMSE | RMSPE | MAE | MAPE | RMSE | RMSPE | MAE | MAPE |
| MLP | 0.0776 | 4.8614 | 0.0673 | 3.3082 | 0.0783 | 2.8201 | 0.0723 | 2.9344 |
| DLSTM | 0.9945 | 15.1296 | 0.9094 | 14.4640 | 0.8432 | 1.3172 | 0.0651 | 2.1788 |
| EyeLSTM | 0.0724 | 4.6658 | 0.0514 | 2.3539 | 0.0565 | 14.2416 | 0.8054 | 15.7320 |

| Algorithm | ClifBar | | | | MotorRolling | | | |
|---|---|---|---|---|---|---|---|---|
| | RMSE | RMSPE | MAE | MAPE | RMSE | RMSPE | MAE | MAPE |
| MLP | 0.1086 | 12.8747 | 0.0995 | 9.0788 | 0.1440 | 28.8403 | 0.1384 | 15.9571 |
| DLSTM | 0.1572 | 16.3077 | 0.1066 | 10.6310 | 0.4237 | 22.4523 | 0.4108 | 30.0869 |
| EyeLSTM | 0.1060 | 10.7539 | 0.0646 | 7.9888 | 0.0526 | 22.2487 | 0.1056 | 13.4955 |

It can be seen from Table 1 that the EyeLSTM algorithm can obtain better fusion results on most datasets.

Combined with Fig.8-9 and Table 1, it can be seen that the EyeLSTM algorithm is better than the MLP and DLSTM algorithms, and its fusion value is closer to the label value, that is, the true value of the target. It shows that the EyeLSTM algorithm has high tracking accuracy and good tracking effect on the target, and realizes the accurate tracking of the target under the conditions of partial occlusion, deformation and illumination.

## 5. conclusion

Although the DLSTM algorithm can handle long-term time series series well, it is difficult to handle tracking data and eye movement data with complex structures. In view of this, in this paper, according to the structural characteristics of eye movement data and tracking data, a new eye tracking data fusion algorithm based on long and short-term memory network is designed. The algorithm first combines CNN and LSTM network, which not only solves the problem that DLSTM algorithm cannot perform parallel computing and obtains global information, but also overcomes the problem that CNN algorithm has no time relationship. Then, a data fusion strategy is designed, that is, the results obtained by the EyeLSTM network training of the eye movement data and the tracking data are weighted and averaged, so that the fusion result obtained is closer to the real value of the target. Finally, 10 sets of tracking data and 10 sets of eye movement data collected by the KCF algorithm and the eye tracker were used to simulate the EyeLSTM, MLP, and DLSTM algorithms. The experimental results show that the fusion result of the EyeLSTM algorithm is closer to the real value of the target, and its fusion results are better in the four evaluation indicators of RMSE, RMSPE, MAE, and MAPE.

**Acknowledges**

The authors would like to thank Professor Jianquan Li for proof-reading the manuscript and the reviewers for their valuable time. This research is supported by the National Natural Science Foundation of China (11971281).